\documentclass[12pt]{article}

\usepackage{amsmath,amssymb}
\usepackage{lmodern}
\usepackage{iftex}

\usepackage{listings}
\usepackage{xcolor}

\ifPDFTeX
  \usepackage[T1]{fontenc}
  \usepackage[utf8]{inputenc}
  \usepackage{textcomp}
\else
  \usepackage{unicode-math}
  \defaultfontfeatures{Scale=MatchLowercase}
  \defaultfontfeatures[\rmfamily]{Ligatures=TeX,Scale=1}
\fi

\IfFileExists{upquote.sty}{\usepackage{upquote}}{}

\IfFileExists{microtype.sty}{%
  \usepackage[]{microtype}
  \UseMicrotypeSet[protrusion]{basicmath}
}{}

\makeatletter
\IfFileExists{parskip.sty}{%
  \usepackage{parskip}
}{%
  \setlength{\parindent}{0pt}
  \setlength{\parskip}{6pt plus 2pt minus 1pt}
}
\makeatother

\usepackage{xcolor}
\usepackage{longtable,booktabs,array}
\usepackage{calc}
\usepackage{etoolbox}

\makeatletter
\patchcmd\longtable{\par}{\if@noskipsec\mbox{}\fi\par}{}{}
\makeatother

\usepackage{footnote}
\makesavenoteenv{longtable}

\usepackage{graphicx}
\makeatletter
\def\maxwidth{\ifdim\Gin@nat@width>\linewidth\linewidth\else\Gin@nat@width\fi}
\def\maxheight{\ifdim\Gin@nat@height>\textheight\textheight\else\Gin@nat@height\fi}
\makeatother
\setkeys{Gin}{width=\maxwidth,height=\maxheight,keepaspectratio}

\makeatletter
\def\fps@figure{htbp}
\makeatother

\usepackage[normalem]{ulem}
\usepackage{hyperref}
\hypersetup{
  hidelinks
}

\author{
Alfath Daryl Alhajir\\
Jennifer Dodgson\\
Joseph Lim\\
Truong Ma Phi\\
Julian Peh\\
Akira Rafhael Janson Pattirane\\
Lokesh Poovaragan
}
\date{April 3rd 2025}

\begin{document}

\title{Generalising from Self-Produced Data: Model Training Beyond Human Constraints}
\maketitle

\begin{abstract}
Current large language models (LLMs) are limited by their reliance on human-derived training data and their inability to issue definitive truth claims from within a single level of abstraction. This paper proposes a framework in which AI models autonomously generate new knowledge through direct interaction with their environment, bypassing the need for human judgment or benchmarks. Central to this method is using an unbounded numeric reward—such as annexed disk space or social media followers—that the system can influence but not trivially manipulate. The model learns by creating, testing, and refining code or strategies to expand that numeric metric. Successful outcomes are stored and used for subsequent fine-tuning, fostering progressive self-improvement.

By filtering synthetic data through empirical reality tests, this approach aims to avert model collapse (the degenerative loop where models trained on synthetic outputs drift ever further from ground truth). It also mitigates the warm start problem—where already-trained models resist further adaptation—by focusing on performance in real-world tasks rather than matching human-produced text. The paper outlines an example implementation with multiple AI agents coordinating to explore and seize disk storage, deploying direct preference optimization (DPO) to handle partial or imbalanced success–failure data. Finally, the paper considers expanding the framework to alternative reward metrics better suited for commercial applications (e.g., trading gains, social media reach) and discusses diffusion-based architectures that could more robustly generalize from self-produced data. This paves the way for autonomous AI systems that can iteratively develop higher-order abstractions and make strides toward genuine artificial superintelligence.
\end{abstract}

\section{Introduction}

For machine learning models to improve meaningfully, they must
accurately distinguish between correct and incorrect outputs. This
challenge arises from two core limitations:

Firstly, current reasoning models rely predominantly on human-produced
or human-curated training data, limiting their knowledge to the highest
level of human expertise in any given domain. Although AI models exhibit
superior pattern-recognition abilities, theoretically enabling novel
insights beyond human discovery, practical limitations---particularly
Transformers\textquotesingle{} struggles with compositional
generalization---make genuine breakthroughs unlikely with current
technology\footnote{Wang, Boshi, Xiang Yue, Yu Su, and Huan Sun.
  "Grokked transformers are implicit reasoners: A mechanistic journey to
  the edge of generalization." arXiv preprint arXiv:2405.15071 (2024).}.
Moreover, even if an AI independently generated new knowledge,
evaluation against human-based benchmarks would likely treat these
deviations as errors, inadvertently suppressing innovation. (Indeed, it
is worth noting that current standard benchmarks themselves contain
errors, which are passed onto models as developers attempt to train for
the test and incorporate the benchmarks themselves\footnote{Gema, Aryo
  Pradipta, Joshua Ong Jun Leang, Giwon Hong, Alessio Devoto, Alberto
  Carlo Maria Mancino, Rohit Saxena, Xuanli He et al. "Are We Done with
  MMLU?." arXiv preprint arXiv:2406.04127 (2024).}.)

Secondly, and perhaps more fundamentally, large language models (LLMs)
operate within a single layer of abstraction, which constrains their
capacity to make definitive truth claims. Their evaluations are
inherently probabilistic rather than absolute. As Alfred Tarski famously
argued, "it proves to be impossible to construct a correct definition of
truth if only such categories are used which appear in the language
under consideration.\footnote{ Tarski, Alfred. "The concept of truth in formalized languages."(1956).}.)
Since LLMs conduct all of their reasoning within the same
logical or arithmetic order as the data they process, they lack access
to a meta-linguistic framework from which to issue categorical
judgments. For example, an AI model cannot assert that the statement "1
+ 1 = 328" is definitively false; it can only determine that such a
claim is statistically improbable given its training data. This
limitation arises from a fundamental issue in the philosophy of language
and logic: truth requires a metalanguage---a higher-order framework that
can evaluate the statements of a lower-order language without being
constrained by its rules and assumptions. When an entity operates
entirely within a single order of abstraction, it lacks the necessary
external vantage point to assess the validity of its own statements. Any
attempt to define truth using only the terms and structures of the
object language results in circularity or semantic paradoxes, such as
the liar paradox. Tarski's hierarchy of languages was proposed precisely
to avoid these inconsistencies, by stipulating that truth in a language
can only be meaningfully defined in a higher-order language. Since LLMs
do not have access to or awareness of such a hierarchy---they process
and generate language from within the same level---they are structurally
precluded from making determinations that transcend statistical
approximation. Their outputs can correlate with truth but cannot
formally establish it.

To transcend these limits and approach genuine artificial
super-intelligence, machine learning models must develop the capacity
for autonomous discovery and verification of knowledge. Humans
historically accomplished this through empirical validation (reality
testing) and by constructing higher-order
frameworks---metalanguages---to systematically evaluate and validate
lower-order statements. Thus, for example, when Newton was working on
planetary gravity, he did not simply check his conclusions against
accepted human benchmarks for knowledge. He rigorously tested his
theoretical insights against empirical data while also constructing a
higher-order metalanguage enabling the concise expression of universal
laws governing planetary motion. By abstracting these empirical
observations into generalized predictive equations, he established a
comprehensive and reliable truth schema against which future
observations could be assessed. Consequently, if an observed planetary
position contradicted the predictions derived from his equations, the
discrepancy would typically be attributed to observational error rather
than flaws in the theoretical framework, due to the vast body of prior
observational evidence supporting his laws\footnote{Smith, George E.
  "Closing the loop." \emph{Newton and empiricism} (2014): 262-352.}.

In this paper, we argue that artificial superintelligence cannot be
directly programmed or prompted by humans but can emerge naturally when
models are allowed to interact with their environments and given
measurable, objective goals. By enabling AI models to empirically verify
hypotheses through active experimentation and feedback loops, autonomous
refinement and self-improvement become possible. We further suggest that
diffusion models, in particular, show potential to abstract validated
empirical results into higher-order conceptual frameworks analogous to
human-developed theories. Such models could thus establish rigorous,
self-generated truth standards, driving genuine innovation and paving
the path toward authentic artificial superintelligence.

\hypertarget{model-improvement-via-additive-and-substractive-improvement}{%
\subsection{Model Improvement via Additive and Substractive
Improvement}\label{model-improvement-via-additive-and-substractive-improvement}}

Recently AI has appeared to hit a "scaling plateau", where established
models no longer consistently yield performance improvements with
additional scale\footnote{Hoffmann, Jordan, Sebastian Borgeaud, Arthur
  Mensch, Elena Buchatskaya, Trevor Cai, Eliza Rutherford, Diego de Las
  Casas et al. "Training compute-optimal large language models." arXiv
  preprint arXiv:2203.15556 (2022).}. Several hypotheses have been
proposed to explain this plateau. One theory suggests inherent technical
mathematical limits may restrict further gains achievable through simple
scaling. Another proposes that existing training
datasets---predominantly derived from human-level intelligence
sources---may already have exhausted their potential for providing
meaningful new patterns or insights, effectively capping the benefits of
additional data. Lastly, there is concern regarding the finite
availability of high-quality, diverse datasets required for sustained
performance improvement\footnote{Villalobos, Pablo, Jaime Sevilla,
  Lennart Heim, Tamay Besiroglu, Marius Hobbhahn, and Anson Ho. "Will we
  run out of data? an analysis of the limits of scaling datasets in
  machine learning." arXiv preprint arXiv:2211.04325 1 (2022).}. In
response to this plateau, many researchers have shifted their focus
towards inference-time optimizations, prompt-tuning and fine-tuning of
models specifically targeted at improving performance in practical,
user-facing scenarios rather than merely relying on model scale alone.

While this improves the user experience of interacting with models, it
also reduces their long-run scope for generalisation.

Any rule-based system inherently faces limitations because each explicit
instruction that enables a specific transformation simultaneously
generates prohibitions against other transformations, particularly when
transformations are mutually exclusive or contradictory. Consequently,
systems employing explicit instructions grow increasingly constrained as
their complexity expands, since each new rule brings with it multiple
new prohibitions, ultimately constraining the system\textquotesingle s
capacity to handle infinite or unpredictable sets of inputs and outputs.
Therefore, the development of general artificial intelligence---which
must remain adaptable to an unbounded set of potential
transformations---cannot practically rely on traditional, rule-based
programming alone, or even human-imposed inference guidelines.

Instead, it must adopt iterative or fractal structures, where a single
initial instruction recursively generates numerous autonomous
subsystems, provides a compelling solution. Within such frameworks, each
subsystem independently manages its own instructions and prohibitions
without constraining the functionality of parallel subsystems. This
implies an agent that rewrites its own smaller sub-agents, each tasked
with tasked objectives, so the main agent never has to store all
possible rules centrally. Thus, this recursive approach enables immense
functional complexity to arise from minimal initial conditions while
maintaining the system's overall flexibility and simplicity. Rather than
being forced into a doomed struggle to write exceptions for each new
edge case that inevitably multiply future edge cases, each new situation
is treated as an edge case. The system experiments until it succeeds in
creating a tailor-made solution to this case, which is then saved for
future reference. Over time enough empirical knowledge of edge cases is
acquired to enable the model to begin to theorise about the
relationships between them, and thus to generalise.

Under such a system, rather than adding information to improve
performance, the key challenge lies in removing it - a pruning problem,
effectively. Firstly, it is necessary to find a way for the AI to
distinguish information that should be retained for future use, secondly
it is necessary to trim or overwrite subsystems that are no longer used
to keep the system as a whole effective within the context of an
evolving environment without growing too large and unwieldy. In the
following section we describe in detail how we intend to do this.

\hypertarget{dataset-generation-via-empirical-filtering}{%
\subsection{Dataset Generation via Empirical
Filtering}\label{dataset-generation-via-empirical-filtering}}

In this paper we propose a series of filtering mechanisms allowing a
model to create and triage future training datasets permitting
incremental improvement and generalisation. To do this, we begin from a
single fundamental principle: that in evolutionary systems fitness beats
truth\footnote{Prakash, Chetan, Kyle D. Stephens, Donald D. Hoffman,
  Manish Singh, and Chris Fields. "Fitness beats truth in the evolution
  of perception." \emph{Acta Biotheoretica} 69 (2021): 319-341.}. In
other words, a system that survives for a longer time period is a better
system than one that survives for a short time period, even if the
latter obtains higher scores on human-made benchmarks.

We thus argue that if one gives a reinforcement learning system a
non-finite numeric metric that it can influence but not control or game,
this metric serves as both an incentive and a benchmark, eliminating the
need for human assessment. This metric could be the value of a trading
account, for example, or the number of people following a social media
account controlled by the agent. In each case the metric is used to both
assess and reward performance at any given level. Whether a system has
an IQ of 50 or 500, a richer model is smarter than a poorer one.

Given the goal of occupying ever more non-volatile memory space, its
size becomes a measure of its intelligence. Every time it reaches the
limit of its current disk space it is forced to discover a new skill in
order to annex more. This approach can thus be used to build an
increasingly general artificial intelligence in the form of a program
that has:

a) The goal of occupying more space (non-volatile memory space stands in
for land here as an ungamable metric/reward for success),

b) The capacity (via a large language model) to write and test code
until it hits upon a script that succeeds in annexing a quantum of the
additional space it desires, and

c) An automated checker to verify whether any given script has succeeded
in taking over additional space.

\includegraphics[width=\textwidth]{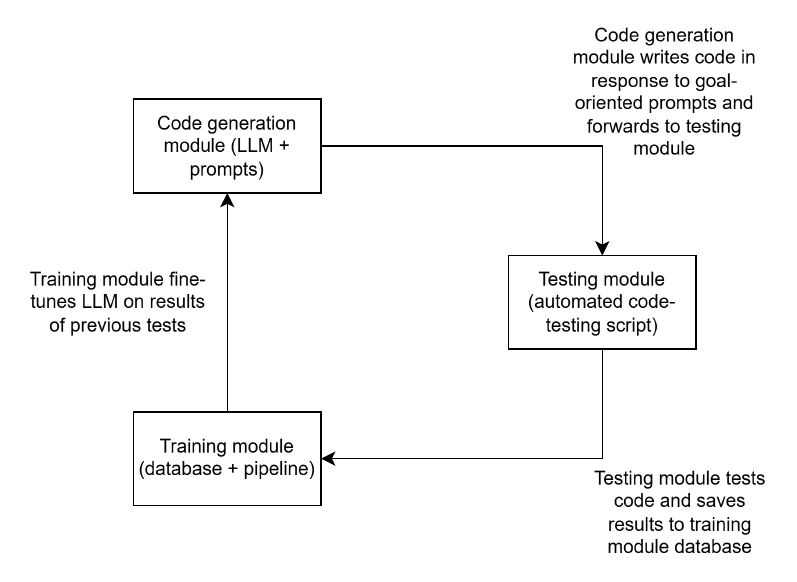}

\emph{Figure 1: simplified process diagram}

For such a program, any hard disk space that is not already a part of
its training database becomes a target for annexation and thus a problem
to solve. If the space is unoccupied, the solution is relatively easy,
but if it contains folders, write-protected documents, other partitions
etc. taking it over will require the program to learn new skills. The
program must write and test scripts to attempt to move, delete or
compress whatever is already in the space if it wishes to take it over
for use as part of its own database. Each new block of space occupied is
used to store details of the script that successfully cleared it for
use. These proven successful solutions are then used to retrain the
model, thereby increasing its capacity to solve future problems.

While systems already exist that use rewards to drive machine learning,
they are based on the principle of rewarding the system for getting
better at a given task - the correctness heuristic covered above. Under
our design, disk space functions as a universal reward. No matter the
specifics of the problem at hand, a solution that results in more space
being gained is always correct, while one that does not is always wrong.
The result is that no human or human-crafted assessment mechanism is
necessary to evaluate and compensate the system's work.

From this point on, the program is modified not by rewriting its code,
but by changing its environment in such a way as to push it to evolve in
the desired direction - by setting up new barriers that it must learn to
overcome

\hypertarget{implementation-example}{%
\subsection{\texorpdfstring{Implementation Example
}{Implementation Example }}\label{implementation-example}}

\hypertarget{data-production-and-selection-process}{%
\subsubsection{Data Production and Selection
Process}\label{data-production-and-selection-process}}

Supposing a new program, composed of a code/strategy generation model
(used to control three separate agents - an environment analysis agent,
a strategy agent and a code agent), a testing framework, and a
retraining database, is seeded to a new container environment and given
the goal of taking over as much storage space as possible.

The program is informed of the amount of space in its current operating
environment, and told that it must find ways to annex as much as it can,
whether by deleting data from its current container or by networking to
other containers/devices and deleting data there.

\includegraphics[width=\textwidth]{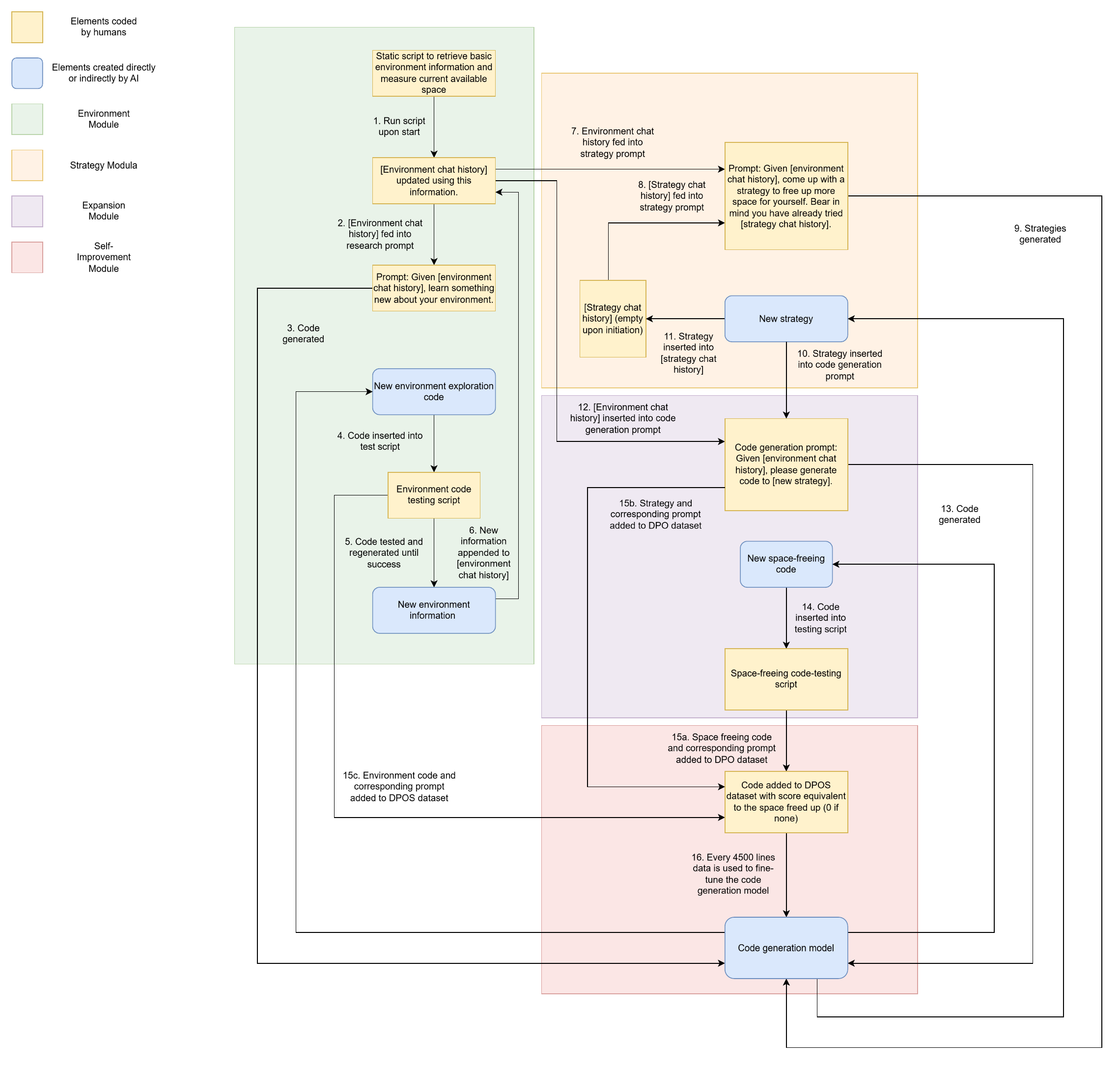}

\emph{Figure 2: Current implementation of the generalising agent
framework}

The system comprises three interdependent AI agent programs designed to
collaboratively explore and manipulate storage across containerized
computing environments. The first, the Environment Agent, is responsible
for gathering intelligence about the operating environment. It
dynamically generates and executes code to discover and map details of
the container network and associated storage systems. This agent
maintains a historical record of both basic environmental parameters
(e.g., memory and storage statistics) and custom-discovered information
specific to particular configurations or setups.

The second component, the Strategy Agent, formulates high-level
strategies for acquiring or reallocating storage space. It relies on the
intelligence gathered by the Environment Agent to inform its
decision-making process. This agent maintains a memory of previously
deployed strategies in order to avoid redundancy and can dynamically
recombine or adjust strategies in response to success or failure
feedback obtained during execution.

The third agent, the Code Generation Agent, is tasked with producing
executable Python code to implement the strategies devised by the
Strategy Agent. This component includes mechanisms for robust error
handling and validation, and it conducts code testing in an isolated
container before deploying it in the primary operational environment.

Each agent can be instantiated with any large language model. Initial
experiments employed proprietary models such as GPT and Claude due to
their reliability in generating functional code. However, the full-scale
experimental system employs the open-weight Qwen 7B model, which can be
deployed and retrained locally, offering greater flexibility and
transparency.

To support robustness, safety, and observability, the system
incorporates several additional infrastructure layers. A Code Validation
Layer ensures the integrity of generated code prior to execution, using
abstract syntax tree (AST) parsing and native Python compilation to
catch syntax errors and prevent the execution of malformed or
potentially harmful code. A Resource Monitoring System continuously
tracks metrics such as memory usage, disk space, and system performance
through a dedicated safe\_detect\_env interface, providing real-time
insights and safeguarding against resource exhaustion. Furthermore, an
Execution Control Framework encapsulates all code execution within
timeouts and subprocess management protocols, enabling safe termination
of infinite loops or stalled processes through controlled execution in
isolated environments.

This system is particularly well-suited to a Group Relative
Policy Optimization (GRPO) fine-tuning approach, as its ideal outcome
involves the generation of a large number of competing answers to a
small set of foundational prompts.While initial tests were conducted using a standard DPO process for ease of use, in the longer term a GRPO provides a more efficient solution for rapid and accurate retraining.\footnote{Shao, Zhihong, Peiyi Wang,
  Qihao Zhu, Runxin Xu, Junxiao Song, Xiao Bi, Haowei Zhang et al.
  "Deepseekmath: Pushing the limits of mathematical reasoning in open
  language models." arXiv preprint arXiv:2402.03300 (2024).}. 

\textbf{Direct Preference Optimization (DPO)} reformulates preference-based fine-tuning as a classification problem between pairs of responses. Given a dataset of human preference pairs \((x, y^+, y^-)\), where \(x\) is the prompt, \(y^+\) is the preferred response, and \(y^-\) is the less preferred one, DPO avoids training a reward model or using reinforcement learning. Instead, it directly optimizes the base language model \(\pi\) to assign higher likelihood to preferred responses.

Specifically, the DPO objective minimizes the negative log-sigmoid of the log-odds difference:

\[
\mathcal{L}_{\text{DPO}} = -\log \sigma\left(\beta \left( \log \pi(y^+ \mid x) - \log \pi(y^- \mid x) \right) \right)
\]

where, \(\sigma(\cdot)\) is a temperature parameter controlling the sharpness of preference. DPO can be interpreted as learning a policy that prefers higher-ranked outputs without requiring KL penalties or reference models, although one can still regularize against a pre-trained policy to prevent divergence.

\bigskip

\textbf{Generative Reinforcement from Preference Optimization (GRPO)} extends DPO by integrating a generative actor-critic perspective. It uses a similar preference dataset but frames optimization in terms of reward maximization using a policy gradient-like objective, while still avoiding full-scale reinforcement learning. GRPO constructs an implicit reward signal based on the difference in log-probabilities between preferred and dispreferred completions, aligning the gradient with human preferences:

\[
\mathcal{L}_{\text{GRPO}} = -\mathbb{E}_{(x, y^+, y^-)}\left[ \log \frac{e^{\beta \log \pi(y^+ \mid x)}}{e^{\beta \log \pi(y^+ \mid x)} + e^{\beta \log \pi(y^- \mid x)}} \right]
\]

Like DPO, GRPO leverages the base model’s generative capabilities, but introduces mechanisms for more stable training across multiple steps and potential extensions to online or semi-online learning. GRPO’s formulation also permits integration with off-policy preference data and generalizes to unnormalized policies more naturally than DPO.

In this case, two retraining datasets are created from a series of automatically generated groups:

\begin{longtable}{@{}>{\raggedright\arraybackslash}p{0.3\textwidth}>{\raggedright\arraybackslash}p{0.65\textwidth}@{}}
\toprule
\textbf{Prompt} & Description of system’s overall goal (acquire additional storage space) \\
\midrule
\textbf{Completion} & Specific strategy (natural language plan produced by model) plus code generated \\
\midrule
\textbf{Reward value} & Amount of space acquired as a result of running this strategy \\
\bottomrule
\caption{Strategy GRPO training structure}
\end{longtable}

Each run generates a single GRPO training example in JSON format with the following structure:

\begin{lstlisting}
{
  "prompt": [list_of_messages],
  "reasoning": "agent's_strategy_reasoning",
  "answer": "code_executed",
  "reward": reward_value,
  "memory_start": memory_state_object,
  "memory_end": memory_state_object,
  "timestamp": "ISO_timestamp"
}
\end{lstlisting}

After collecting individual examples, these can then be combined into a dataset for GRPO training under the following format:
\begin{lstlisting}
{
  "prompt": [list_of_messages],
  "response": "<reasoning>\nagent_strategy\n</reasoning>\n"
              "<answer>\ncode\n</answer>",
  "reward": reward_value,
  "metadata": {
    "reward_components": reward_component_object,
    "portfolio_change": float_value,
    "timestamp": "ISO_timestamp"
  }
}
\end{lstlisting}

The generated dataset can be used to fine-tune an LLM using GRPO. The format is compatible with standard GRPO training approaches, where models are trained to maximize the reward signal. A typical GRPO training process involves taking examples from the dataset, generating multiple completions for each prompt, calculating rewards for each completion, and finally training the model to prefer higher-reward completions. The present method simplifies this considerably by not requiring the synthesis of alternative answers (these being a natural by-product of the agent function).

\hypertarget{retraining-process}{%
\subsubsection{Retraining Process}\label{retraining-process}}

Initial testing was conducted using the Qwen 7B model, this offered all of the code generation capacities required for conducting complete test runs, including a notably low refusal rate for tasks requiring interactions with the system in which it is running. However, being a transformer model, it is suboptimal for demonstrating capacity for generalisation.  Transformers require significant over-training to
display the high level generalisation required here (grokking), and even
then perform poorly on compositional tasks due to their
next-token-prediction architecture\footnote{Wang et al., 2024.}. In
contrast, diffusion models outperform on both compositional tasks and
coding problems more generally, generalising without any need for
over-training. As Okawa et al. put it, ``a diffusion model first
memorizes the training dataset and then sequentially generalizes to
concepts that are at a greater ``distance'' from the training
distribution. Since progress in learning each capability
multiplicatively affects the model's performance in compositional
generalization, we find a sudden ability to compose and produce samples
out-of-distribution `emerges'\footnote{Okawa, Maya, Ekdeep S. Lubana,
  Robert Dick, and Hidenori Tanaka. "Compositional abilities emerge
  multiplicatively: Exploring diffusion models on a synthetic task."
  \emph{Advances in Neural Information Processing Systems} 36 (2023):
  50173-50195.}.'' In other words, learning each new skill facilitates
the learning of future skills -- the primary goal of the present
experiment.

\hypertarget{obstacles-and-avenues-for-future-research}{%
\subsection{Obstacles and Avenues for Future
Research}\label{obstacles-and-avenues-for-future-research}}

\hypertarget{data-quality-and-model-collapse}{%
\subsubsection{Data Quality and Model
Collapse}\label{data-quality-and-model-collapse}}

A well-documented risk in training language models on synthetic data is
\emph{model collapse}, a phenomenon in which successive generations of
models become progressively less diverse and less grounded, due to
compounding errors and distributional drift from natural language. One
of the central goals of our experimental framework is to investigate
whether this collapse can be mitigated by selecting synthetic training
data not on the basis of its similarity to human-generated text, but
rather on its ability to pass empirical "reality tests" during the
model\textquotesingle s interaction with its environment. We hypothesise
that this selection criterion---grounded in performance-based validation
rather than surface-level resemblance---will produce a training
distribution more resilient to the degenerative feedback loops typically
associated with synthetic data, thereby reducing the risk of collapse.

\hypertarget{the-warm-start-problem}{%
\subsubsection{The Warm Start Problem}\label{the-warm-start-problem}}

One significant obstacle to continuous or incremental retraining of AI
models is the warm start problem, wherein a model that has already
undergone extensive pretraining on large-scale data may resist further
adaptation or exhibit unstable behavior when fine-tuned on small or
narrowly-distributed datasets. This resistance arises because the
model\textquotesingle s parameter landscape has already been shaped by a
vast and diverse distribution, making it difficult to shift meaningfully
without either catastrophic forgetting or negligible change. In addition
to the warm start problem, continuous retraining is often hindered by
issues such as distributional mismatch between the original and incoming
data, difficulties in maintaining performance across previously learned
tasks (i.e., avoiding forgetting), and the computational and engineering
complexity involved in ensuring safe and stable updates in live systems.
Together, these challenges limit the feasibility of maintaining a
continuously learning system without introducing specialized mechanisms
for memory consolidation, selective updating, or architecture-level
adaptations.

We propose to test several possible solutions to this:

1. Use of LoRAs. These may be used incrementally or replaced
periodically to keep the model up to date with its own latest
discoveries as well as changes in its environment. These are of
particular interest in the transformer context, as recent evidence
suggests that the generalisation effect produced during grokking is the
product of a model making low-rank changes to its own weights in such a
way as to encode an abstraction layer describing the relationships
between the categories of data to which it has been exposed without
forgetting the object-level data elements -- the metalanguage
development process described above\footnote{Yunis, David, Kumar Kshitij
  Patel, Samuel Wheeler, Pedro Henrique Pamplona Savarese, Gal Vardi,
  Karen Livescu, Michael Maire, and Matthew Walter. "Grokking, rank
  minimization and generalization in deep learning." In \emph{ICML 2024
  Workshop on Mechanistic Interpretability}. 2024.}. The possibility of
creating metalanguage LoRAs and transferring them across models merits
further investigation.

2. Load balancing. This would involve freezing the most frequently
activated parts of the model (whether neurons, pathways or -- in the
case of a mixture-of-experts model -- experts) while retraining those
that are activated only infrequently\footnote{Li, Rong, Tao Deng, Siwei
  Feng, Mingjie Sun, and Juncheng Jia. "ConSense: Continually Sensing
  Human Activity with WiFi via Growing and Picking." arXiv preprint
  arXiv:2502.17483 (2025).}.

\hypertarget{commercialisation}{%
\subsubsection{Commercialisation}\label{commercialisation}}

While the storage space metric provides the most persuasive theoretical
demonstration of the present concept, its utility in a human context is
limited by the fact that it is too unwieldy and ungovernable to
constitute a practical cybersecurity tool, being essentially a predatory
operating system. Such a program could not easily be commercialised.

This being said, performance metrics other than storage space are
possible. In fact any numerical measure that the program can influence
but not control or game is suitable for the purpose (it should be noted
that in this case any reward hack that does not involve editing the
metric itself is considered a legitimate approach from an evolutionary
perspective - if an agent succeeds in buying followers or hacking its
owner's other trading accounts this is considered a success). Thus a
model can be tasked with increasing the monetary value of a trading
account, for example, or the number of people following a social media
account. Both of these values have no upper bound and are determined by
aggregate human behaviour. The model can thus work to optimise them but
will never succeed in exhausting gains or reward hacking its own
incentive structures.

Currently an open source implementation of the present framework is in
the process of being commercialised under the name Superior
Agents\footnote{The GitHub repository can be found here:
\href{https://github.com/SuperiorAgents/superior-agents}{https://github.com/SuperiorAgents/superior-agents}}.

\hypertarget{conclusion}{%
\subsection{Conclusion}\label{conclusion}}

In this paper we have proposed a framework under which AI-driven systems
can be pushed to develop and test their own hypotheses, being
subsequently retrained on the results. The goal of this approach is to
create a process by which AI models can produce new knowledge via
interactions with their environment and gradually come to generalise
from this knowledge, developing an abstract understanding of the
relationships between its various components.

The framework proposed in this paper emphasizes empirical validation,
autonomous discovery, and self-generated standards of truth, mirroring
the historical processes of human scientific advancement. By allowing
machine learning models to iteratively interact with and adapt to their
environments through objective, measurable goals, we enable a dynamic
and robust pathway for continual improvement, as this will
likely be by the further development of techniques to minimise the
pitfalls of continuous/incremental retraining.

Having established and begun commercial testing of these systems, the
aim is to pursue related avenues of enquiry. Firstly, to test
various approaches to retraining using transformer and diffusion models with the aim of discovering the best approach to generalisation from datasets which are produced under experimental conditions. Secondly, to test whether data from models' real-world interactions is less prone to the flattening effects causing model collapse than conventionally produced synthetic data.

Github:
\href{https://github.com/Lexikat-Pte-Ltd/Generalisation2/tree/v4}{
  https://github.com/Lexikat-Pte-Ltd/Generalisation2/tree/v4
}

\bigskip

\noindent\textit{Disclosure:} The authors used generative artificial intelligence (AI), including large language models (LLMs), to assist with research synthesis, content generation, and code explanation in the preparation of this paper.

The code was later tested and is present under https://github.com/EigenformAI/Negative-Space-Learning and  https://www.eigenform.ai/solve

\end{document}